\documentclass[runningheads]{llncs}
\usepackage{graphicx}
\usepackage[dvipsnames,table]{xcolor}
\usepackage{tikz}
\usepackage{comment}
\usepackage{amsmath,amssymb} 
\usepackage{multirow}
\usepackage[pagebackref,breaklinks,colorlinks]{hyperref}

\usepackage[accsupp]{axessibility}  

\usepackage[width=122mm,left=12mm,paperwidth=146mm,height=193mm,top=12mm,paperheight=217mm]{geometry}

\DeclareMathOperator*{\agg}{agg}
\newcommand{\OURS}{Texturify}

\begin{document}
\pagestyle{headings}
\mainmatter

\title{\OURS{}: Generating Textures on \\3D Shape Surfaces}

\makeatother
\newcommand{\MATTHIAS}[1]{{\emph{\textcolor{red}{\textbf{Matthias:~#1}}}}}
\newcommand{\ANGIE}[1]{{\emph{\textcolor{blue}{Angie: #1}}}}
\newcommand{\JT}[1]{{\emph{\textcolor{magenta}{Justus: #1}}}}
\newcommand{\YAWAR}[1]{{\emph{\textcolor{ForestGreen}{Yawar:~#1}}}}
\newcommand{\FANGCHANG}[1]{{\emph{\textcolor{brown}{Fangchang:~#1}}}}
\newcommand{\QI}[1]{{\emph{\textcolor{brown}{Qi:~#1}}}}
\newcommand{\TODO}[1]{{\emph{\textcolor{red}{TODO: #1}}}}

\definecolor{LightBlue}{rgb}{0.827,0.870,0.952}

\titlerunning{Texurify}

\author{Yawar Siddiqui\inst{1} \and
Justus Thies\inst{2} \and
Fangchang Ma\inst{3} \and 
Qi Shan\inst{3} \and \\
Matthias Nie{\ss}ner\inst{1} \and
Angela Dai\inst{1}
}

\authorrunning{Y. Siddiqui et al.}

\institute{Technical University of Munich \and
Max Planck Institute for Intelligent Systems \and
Apple}

\maketitle

\vspace{-1cm}
\begin{figure}
\centering
\includegraphics[width=\linewidth]{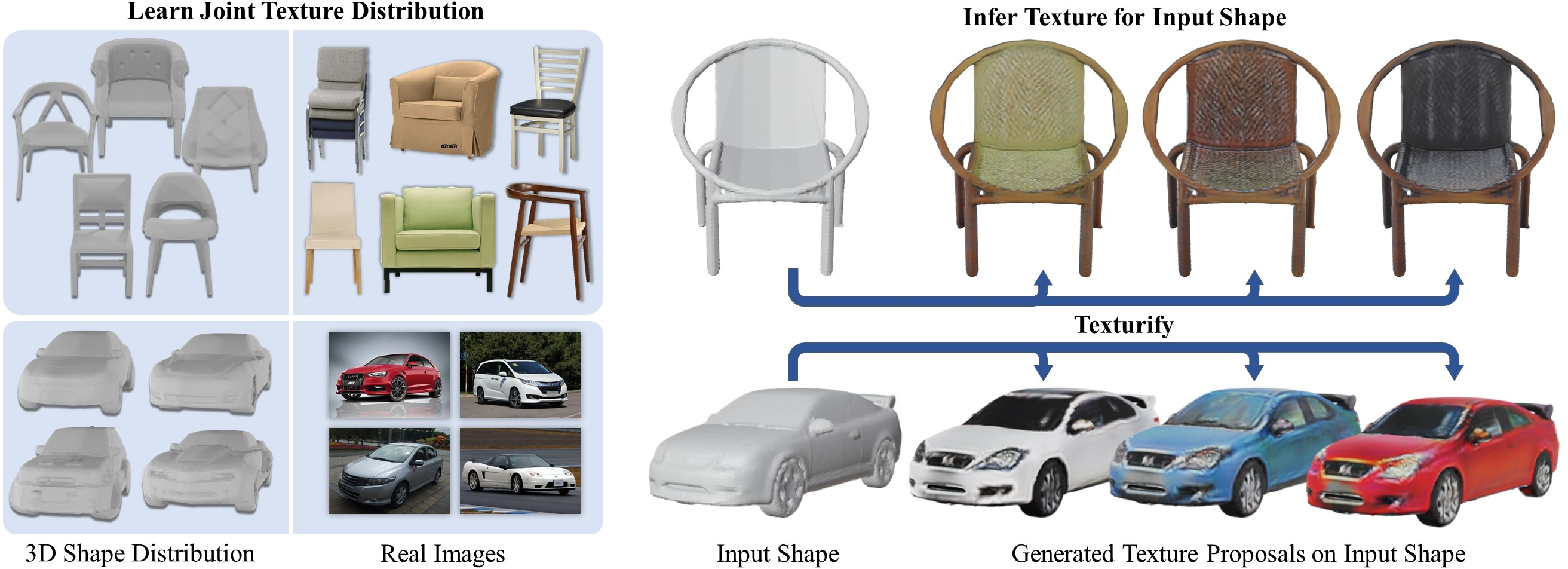}
\vspace{-0.75cm}
\caption{\textit{\OURS{}} learns to generate geometry-aware textures for untextured collections of 3D objects. Our method produces textures that when rendered to various 2D image views, match the distribution of real image observations. 
\textit{\OURS{}} enables training from only a collection of images and a collection of untextured shapes, which are both often available, without requiring any explicit 3D color supervision or shape-image correspondence.
Textures are created directly on the surface of a given 3D shape, enabling generation of high-quality, compelling textured 3D shapes.
}
\label{fig:example}
\end{figure}
\vspace{-1.0cm}
\begin{abstract}
Texture cues on 3D objects are key to compelling visual representations, with the possibility to create high visual fidelity with inherent spatial consistency across different views.
Since the availability of textured 3D shapes remains very limited, learning a 3D-supervised data-driven method that predicts a texture based on the 3D input is very challenging.
We thus propose \textit{\OURS{}}, a GAN-based method that leverages a 3D shape dataset of an object class and learns to reproduce the distribution of appearances observed in real images by generating high-quality textures.
In particular, our method does not require any 3D color supervision or correspondence between shape geometry and images to learn the texturing of 3D objects.
\textit{\OURS{}} operates directly on the surface of the 3D objects by introducing face convolutional operators on a hierarchical 4-RoSy parameterization to generate plausible object-specific textures.
Employing differentiable rendering and adversarial losses that critique individual views and consistency across views, we effectively learn the high-quality surface texturing distribution from real-world images.
Experiments on car and chair shape collections show that our approach outperforms state of the art by an average of 22\% in \textit{FID} score.
%

\end{abstract}

\section{Introduction}
3D content is central to many application areas, including content creation for visual consumption in films, games, and mixed reality scenarios. 
Recent years have seen remarkable progress in modeling 3D geometry~\cite{chibane2020implicit,dai2020sg,mescheder2019occupancy,deepsdf,peng2020convolutional,siddiqui2021retrievalfuse,azinovic2021neural}, achieving significant improvements on geometric fidelity, driven by new generative learning approaches on large-scale 3D shape datasets~\cite{shapenet2015,fu20213d,pix3d}.
While strong promise has been shown in modeling geometry, generating fully textured 3D objects remains a challenge that is less explored.
As such, textured 3D content generation still demands tedious manual efforts.
A notable challenge in learning to automatically generate textured 3D content is the lack of high-quality textured 3D data.
Large-scale shape datasets such as ShapeNet~\cite{shapenet2015} have helped to drive the success of 3D geometric shape modeling, but tend to contain simplistic and often uniform textures associated with the objects.
Furthermore, existing texture generation approaches primarily follow popular generative geometric representations that define surfaces implicitly over a volume in space~\cite{dai2021spsg,oechsle2019texture,sun2018im2avatar}, which results in inefficient learning and tends to produce blurry results, since  textures are only well-defined on geometric surfaces.
We propose \textit{\OURS{}} to address these challenges in the task of automatic texture generation for 3D shape collections.
That is, for a given shape geometry, \textit{\OURS{}} learns to automatically generate a variety of different textures on the shape when sampling from a latent texture space.
Instead of relying on supervision from 3D textured objects, we utilize only a set of images along with a collection of 3D shape geometry from the same class category, without requiring any correspondence between image and geometry nor any semantic part information of the shapes.
We employ differentiable rendering with an adversarial loss to ensure that generated textures on the 3D shapes produce realistic imagery from a variety of views during training.
Rather than generating textures for 3D shapes defined over a volume in space as has been done with implicit representations~\cite{chanmonteiro2020pi-GAN,ChanCVPR20201,Pan2021ASG,SchwarzNEURIPS2020} or volumetric representations~\cite{dai2021spsg,sun2018im2avatar}, we instead propose to tie texture generation directly to the surface of the 3D shape.
We formulate a generative adversarial network, conditioned on the 3D shape geometry and a latent texture code, to operate on the faces of a 4-way rotationally symmetric quad mesh by defining face convolutional operators for texture generation.
In contrast to the common 2D texture parameterization with UV maps, our method enables generating possible shape textures with awareness of 3D structural neighborhood relations and minimal distortion.
We show the effectiveness of \OURS{} in texturing ShapeNet chairs and cars, trained with real-world imagery; our approach outperforms the state of the art by an average of 22\% FID scores.
%

\noindent
In summary, our contributions are:
\begin{itemize}
    \item A generative formulation for texture generation on a 3D shape that learns to create realistic, high-fidelity textures from 2D images and a collection of 3D shape geometry, without requiring any 3D texture supervision.
    \item A surface-based texture generation network that reasons on 3D surface neighborhoods to synthesize textures directly on a mesh surface following the input shape geometry and a latent texture code.
\end{itemize}

\section{Related Works}
\textit{\OURS{}} aims at generating high-quality textures for 3D objects.
Our generative method is trained on distinct sets of 3D shape and real 2D image data.
Related approaches either require aligned 2D/3D datasets to optimize for textures, meshes with surface color, or learn a joint distribution of shape and appearance directly from 2D images.
In contrast, our method predicts textures for existing 3D shapes using a parameterization that operates directly on the surface and convolutions on the 3D meshes.

\smallskip
\noindent \textbf{Texturing via Optimization.}
Texture optimization methods address instance-specific texture generation by iteratively optimizing over certain objective functions requiring aligned 2D/3D data.
Traditional methods~\cite{zhou2014color} employ global optimization for mapping color images onto geometric reconstructions.
To improve robustness against pose misalignments, Adversarial Texture Optimization~\cite{huang2020adversarial} reconstructs high-quality textures from RGBD scans with a patch-based, misalignment-tolerant conditional discriminator, resulting in sharper textures. 
Note that these methods are not able to complete missing texture regions and heavily rely on the provided input images.
Recently, Text2Mesh~\cite{michel2021text2mesh} proposes to optimize for both color and geometric details of an input mesh to conform to a target text using a CLIP-based~\cite{radford2021learning} multi-view loss.

\smallskip
\noindent \textbf{Texture Completion.}
IF-Net-Texture~\cite{chibane2020texture} focuses on texture completion from partial textured scans and completed 3D geometry using IF-Nets~\cite{chibane2020shape}, with a convolutional encoder and an implicit decoder.
This method learns locally-implicit representations and requires 3D supervision for training.
SPSG~\cite{dai2021spsg} proposes a self-supervised approach for training a fully convolutional network to first predict the geometry and then the color, both of which are represented as 3D volumetric grids.
In comparison to these completion methods, our proposed method only requires an uncoupled collection of 3D shapes and 2D images as supervision. 

\smallskip
\noindent \textbf{Retrieval-based Texturing.}
PhotoShape~\cite{park2018photoshape} proposes retrieval-based texturing using a dictionary of high-quality material assets.
Specifically, it classifies the part material of an object in a 2D image and applies the corresponding material to the respective parts of a 3D object.
While this approach is able to generate high-quality renderings, it requires detailed segmentations in the 2D image as well as for the 3D object, and is unable to produce material that is not present in the synthetic material dataset.

\smallskip
\noindent \textbf{Generative Texturing.}
TextureFields~\cite{oechsle2019texture} learns a continuous representation for representing texture information in 3D, parameterized by a fully connected residual network.
While this approach can be trained in an uncoupled setting, it tends to produce blurry or uniform textures (see Sec.~\ref{sec:results}).
The work closest to ours in terms of problem formulation is Learning Texture Generators From Internet Photos (LTG) ~\cite{yu2021learning}, where the texture generation task is formulated as a shape-conditioned StyleGAN~\cite{Karras2019ASG} generator.
This approach requires several different training sets, each containing images and silhouettes from similar viewpoints, and multiple discriminators are employed to enforce correct rendering onto these corresponding viewing angles.
In comparison, our method makes no explicit assumptions on the partitioning of viewpoints.
LTG also makes use of UV parameterization, where the texture atlases come from fixed views around the object, which inevitably results in seams and could struggle for non-convex shapes, whereas our method operates directly on mesh faces and is thus free from these two issues.
Additionally, LTG utilizes only the silhouettes but not surface features for training, but our approach is geometry-aware.

\noindent \textbf{Generative Models for Geometry and Appearance.}
3D-aware image synthesis and automatic generation of 3D models have gained attention recently.
Early approaches use discretized spatial representations such as a 3D voxel grid~\cite{Gadelha3DV2017,HenzlerICCV2019,NguyenICCV2019,NguyenNEURIPS2020,WuNIPS2016,xu2022point,ZhuNEURIPS2018}, with the downside of high memory footprint that grows cubically with resolution.
Scaling up to high-resolution image generation with 3D voxel grids can be challenging, even with sparse volumetric representations~\cite{HaoICCV2021}.
To further alleviate memory consumption, upsampling of synthesized images with 2D convolutional networks has been proposed~\cite{NiemeyerCVPR2021}.
An alternative approach is neural implicit representations~\cite{ChanCVPR20201,Pan2021ASG,SchwarzNEURIPS2020,gafni2021dynamic} which encode both the geometry and texture into a single multi-layer perceptron (MLP) network.
Seminal work along this line of research includes HoloGAN~\cite{NguyenICCV2019}, GIRAFFE~\cite{Niemeyer2021GIRAFFERS}, GRAF~\cite{SchwarzNEURIPS2020}, and PiGAN~\cite{chanmonteiro2020pi-GAN}.
Several concurrent works focus on the generation of high-resolution images with implicit representations, for example, EG3D~\cite{ChanARXIV2021}, StyleNeRF~\cite{GuARXIV2021}, CIPS-3D~\cite{ZhouARXIV2021}, and StyleSDF~\cite{OrEl2021StyleSDFH3}.
These approaches generate high-quality 2D views rather than the textures for given input meshes, which is the focus of this paper.
Additionally, these works generate both the pseudo-geometry and colors in a coupled fashion, resulting in tangled geometry and texture that are difficult to be separated for downstream applications.
More recent work~\cite{pavllo2021learning} generates textured triangle meshes from either random noise latent or 3D semantic layouts, which is also different from our problem formulation which conditions on untextured geometry.

\noindent \textbf{Convolutions on 3D Meshes.}
Several approaches have been proposed for applying convolutions on meshes~\cite{verma2018feastnet,masci2015geodesic,tatarchenko2018tangent,hanocka2019meshcnn,huang2019texturenet}.
For instance, MeshCNN~\cite{hanocka2019meshcnn} proposes a trimesh-based edge convolution operator, demonstrating part segmentation on relatively small or decimated 3D shape meshes, while our approach aims to generate high-fidelity textures on the faces of high-resolution shape meshes.
TangentConv~\cite{tatarchenko2018tangent} proposes tangent convolution, which projects local surface geometry on a tangent plane around every point and applies 2D planar convolutions within these tangent images.
TextureConv~\cite{huang2019texturenet} defines a smooth, consistently oriented domain for surface convolutions based on four-way rotationally symmetric (4-RoSy) fields and demonstrate superior performance compared to TangentConvs for semantic segmentation.
Our approach builds on this TextureConv 4-RoSy parameterization to generate textures on mesh faces.

\vspace{-0.2cm}

\section{Method}

Given a collection of untextured meshes of a  class of objects, our method aims to learn a texture generator using only 2D image collections as supervision. 
We do not assume any correspondence between the shapes and the image collection, except that they should represent the same class category of objects.

\begin{figure}[h]  
    \centering
    \includegraphics[width=0.96\linewidth]{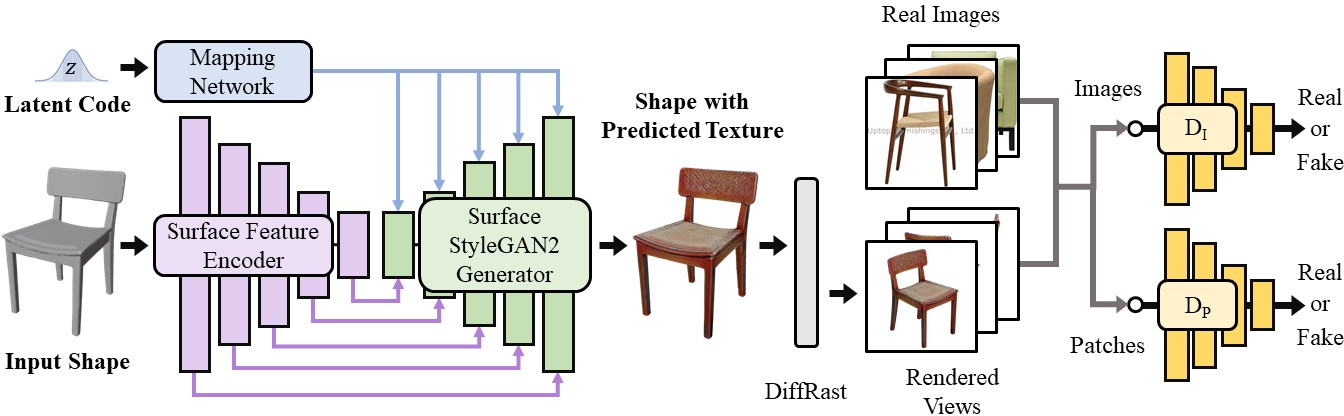}
    \caption{\textit{\OURS{}} Overview. Surface features from an input 3D mesh are encoded through a face convolution-based encoder and decoded through a StyleGAN2-inspired decoder to generate textures directly on the surface of the mesh.
    To ensure that generated textures are realistic, the textured mesh is differentiably rendered from different view points and is critiqued by two discriminators. An image discriminator $\mathrm{D}_I$ operates on full image views from the real or rendered views, while a patch-consistency discriminator $\mathrm{D}_P$ encourages consistency between views by operating on patches coming from a single real view or patches from different views of rendered images.}
    \label{fig:overview}
\end{figure}

\noindent
We propose a generative adversarial framework to tackle this problem (see Fig.~\ref{fig:overview}). 
Given a 3D mesh of an object, and latent texture code, our generator produces textures directly on the mesh surface. 
To this end, we parameterize the mesh with a hierarchy of 4-way rotationally symmetric (4-RoSy) fields from QuadriFlow~\cite{QuadriFlow} of different resolutions.
This parameterization enables minimal distortion without seams and preserving geodesic neighborhoods. 
Thus we define convolution, pooling and unpooling operations that enable processing features on the mesh surface and aggregate features across resolution hierarchy levels. 
Using these operators, we design the generator as a U-Net encoder-decoder network, with the encoder as a ResNet-style feature extractor and the decoder inspired by StyleGAN2~\cite{Karras2020AnalyzingAI}, both modified to work on object surfaces.
We use differentiable rendering to render the 3D meshes with the generated textures and enforce losses to match the appearance distribution of real image observations. 
Specifically, we apply two discriminators against the real image distribution for supervision: 
the first discriminator is inspired by StyleGAN2 on individual rendered images to match the real distribution, while the other encourages global texture consistency on an object through patch discrimination across multiple views.
The pipeline is trained end-to-end using a non-saturating GAN loss with gradient penalty and path length regularization.

\subsection{Parameterization}
\begin{figure}[t]  
\centering
\includegraphics[width=\linewidth]{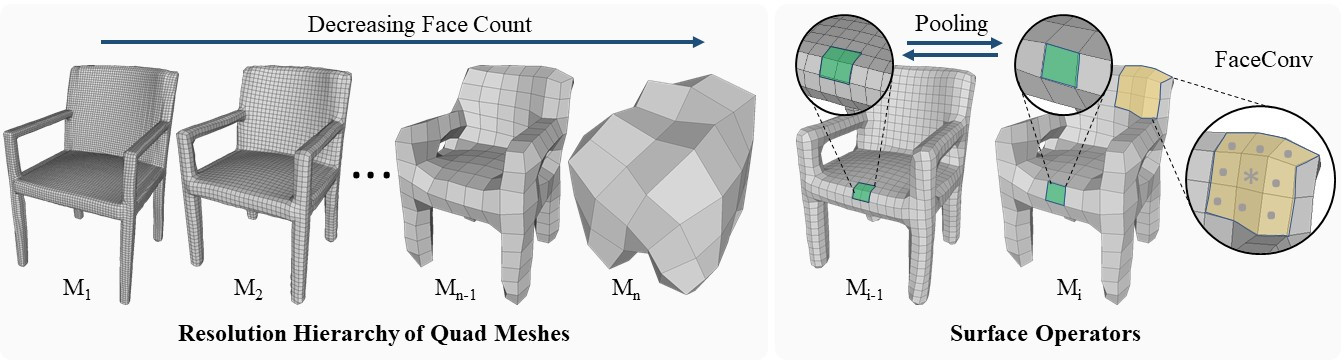}
\caption{\OURS{} generates textures on 4-RoSy parameterized quad meshes, where we define face convolutions with pooling and unpooling operations to operate on a hierarchy of the quad meshes. This enables reasoning about local surface neighborhoods across multiple resolutions to obtain effective global structures as well as fine details.}
\label{fig:hierarchy}
\end{figure}

Since textures are naturally a surface attribute, parameterizing them in 3D space is inefficient and can result in blurry textures. 
Therefore, we aim to generate textures directly on the surface using a surface parameterization. 
One popular way of representing textures on the surface of a mesh is through UV mapping. 
However, generating a consistent UV parameterization for a set of shapes of varying topology is very challenging. 
Furthermore, it can introduce distortions due to flattening as well as seams at surface cuts. 
Seams in particular make  learning using neighborhood-dependent operators (e.g., convolutions) difficult, since neighboring features in UV space might not necessarily be neighbors in the geodesic space.
To avoid these issues, we instead generate surface texture on the four-fold rotationally symmetric (4-RoSy) field parameterization from Quadriflow~\cite{QuadriFlow}, a method to remesh triangle meshes as quad meshes. 
A 4-RoSy field is a set of tangent directions associated with a vertex where the neighboring directions are parallel to each other by rotating one of them around their surface normals by a multiple of ${90^\circ}$. 
This can be realized as a quad-mesh, without seams and with minimal distortions, near-regular faces, and preservation of geodesic neighborhoods, making it very suitable for convolutional feature extraction (as shown in Fig.~\ref{fig:hierarchy}). 
To facilitate a hierarchical processing of features on the surface of the mesh, we precompute this 4-RoSy field representation of the mesh $\mathcal{M}$ at multiple resolution levels to obtain quad meshes $\mathrm{M}_1, \mathrm{M}_2, .., \mathrm{M}_n$ with face count $\frac{|M_1|}{4^{l-1}}$, with $|M_1|$ being the face count at the finest level (leftmost in Fig.~\ref{fig:hierarchy}).

\subsection{Surface Operators}
Given a 4-RoSy parameterized quad-mesh, we process features directly on its surface by defining convolutions on the faces. 
A face convolution operates on a face with feature $\boldsymbol{\mathrm{x}}_i$ and its neighboring faces' features $\mathcal{N}_i = [\boldsymbol{\mathrm{y}}_{1}, \boldsymbol{\mathrm{y}}_{2}, ...]$ by:
\begin{equation}
    \mathrm{FaceConv}(\boldsymbol{\mathrm{x}}_i, \mathcal{N}_i) = \boldsymbol{\mathrm{w}}^{T}_{0}\boldsymbol{\mathrm{x}}_i + \sum_{j=1}^{|\mathcal{N}_i|} \boldsymbol{\mathrm{w}}^{T}_{j}\boldsymbol{\mathrm{y}}_j + \boldsymbol{\mathrm{b}}
\end{equation}
with $\boldsymbol{\mathrm{x}}_i \in \mathbb{R}^{C_0}$, $\boldsymbol{\mathrm{y}}_j \in \mathbb{R}^{C_0}$, learnable parameters $\boldsymbol{\mathrm{w}} \in \mathbb{R}^{C_0 \times C_1}$, $\boldsymbol{\mathrm{b}} \in \mathbb{R}^{C_1}$, where ${C_0}$ and ${C_1}$ are input and output feature channels respectively. 
We use a fixed face neighborhood size  $|\mathcal{N}_i|=8$, since the vast majority of the faces in the quad mesh have a neighborhood of 8 faces, with very few singularities from Quadriflow remeshing.
In the rare case of singularities (see suppl. doc.), we zero-pad the faces so that the number of neighbors is always 8. 
Additionally, neighbors are ordered anticlockwise around the face normal, with the face having the smallest Cartesian coordinates (based on $x$, then $y$, then $z$) as the first face.

For aggregating features across mesh resolution hierarchy levels, we define inter-hierarchy pooling and unpooling operators.
The pooled features $\boldsymbol{\mathrm{x}}_{l+1}^{j}$ are given as $\boldsymbol{\mathrm{x}}_{l+1}^{j} = \agg
    \left( \left\{ \boldsymbol{\mathrm{x}}_{l}^{i} : {i \in \mathrm{F}^j_{l+1}}\right\} \right)$, 
where $\mathrm{F}_{l+1}^j$ defines the set of face indices of the finer layer $l$ which are nearest to the $j$-th face of the coarser layer ${l+1}$ in terms of a chamfer distance and with `$\agg$' as an aggregation operator.
The unpooled features are computed as:
    $\boldsymbol{\mathrm{x}}_{l}^{j} = \boldsymbol{\mathrm{x}}_{l+1}^{\hat{\mathrm{F}}_{l}^j}$,
where $\hat{\mathrm{F}}_{l}^j$ defines correspondence of the $j$-th face of the fine layer to the coarse layer (in terms of minimal chamfer distance).

\subsection{Surface Texture GAN Framework}
With our hierarchical surface parameterization and surface features operators, we design a GAN framework that generates colors on the mesh surface that can be trained using only a collection of images and untextured shapes without any explicit 3D texture supervision.
Our generator takes a U-shaped encoder-decoder architecture. 
Face normals and mean curvature are used as input features to the network. 
The encoder is then designed to extract features from the mesh surface at different resolution levels. 
These features are processed through a series of FaceResNet blocks (ResNet blocks with FaceConv instead of Conv2D) and inter-hierarchy pooling layers as defined above. 
Features extracted at each level of the hierarchy are then passed to the appropriate level of the decoder through U-Net skip connections. 
This multi-resolution understanding is essential towards generating compelling textures on a 3D shape, as the deeper features at coarse hierarchy levels enable reasoning about global shape and textural structure, with finer hierarchy levels allowing for generation of coherent local detail.

The decoder is inspired by the StyleGAN2~\cite{Karras2020AnalyzingAI} generator, which has proven to be a stable and efficient generative model in 2D domain. 
We thus use a mapping network to map a latent code to a style code. 
Additionally, we upsample (via inter-hierarchy unpooling) and sum up RGB contributions at different mesh hierarchy levels, analogous to StyleGAN2 upsampling and summation at different image resolutions.
Instead of the style-modulated Conv2D operators of StyleGAN2 style blocks, we use style-modulated FaceConvs.
In contrast to the StyleGAN2 setting, where the generated image has a fixed structure, we aim to generate textures on varying input 3D shape geometries. Therefore, we further condition the style blocks on surface features extracted by the mesh encoder. 
This is achieved by concatenating the features generated by the decoder at each hierarchy level with the encoder features from the same level. 
The decoder outputs face colors for the highest resolution mesh ($l=1$) representing the texture.

\begin{figure}[t]  
    \centering
    \includegraphics[width=\linewidth]{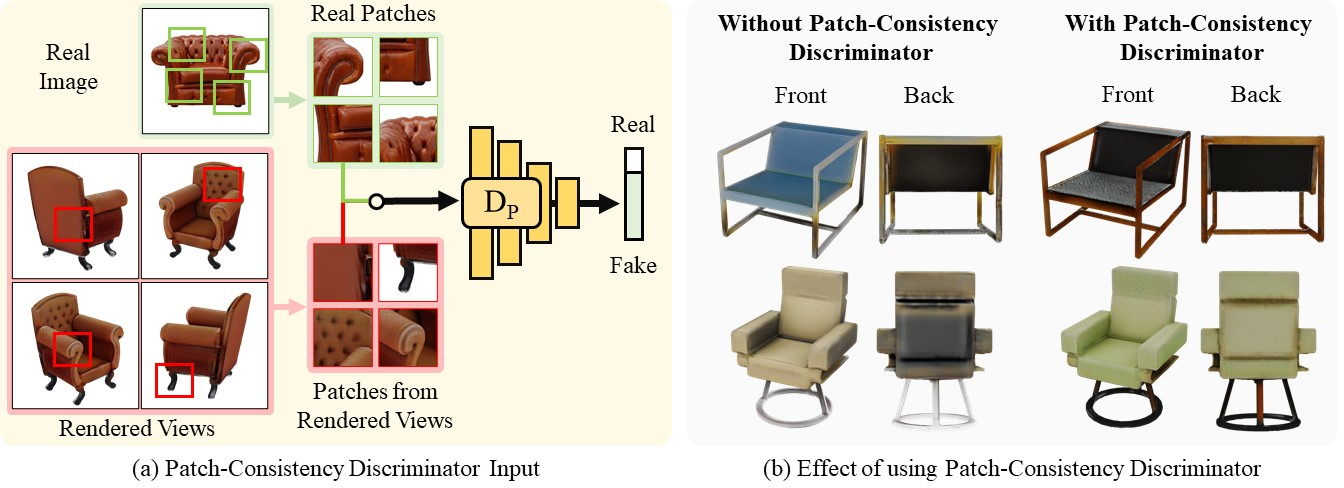}
    \caption{
    The patch-consistency discriminator encourages global consistency in generated shape textures across multiple views.
    (a) While for real image data we are only considering patches from the same view (since the 2D image dataset does not contain multi-view data), we use patches from multiple views in the scenario of generated images. (b) Without patch-consistency discriminator rendered texture can end up having inconsistent styles across viewpoints. Using the patch consistency discriminator prevents this issue.  
    }
    \label{fig:method_patchdisc}
    \vspace{-0.4cm}
\end{figure}

To enable training using only 2D image collections for texture supervision, the resulting textured mesh is rendered as an image from multiple viewpoints using a rasterization-based differentiable renderer, Nvdiffrast~\cite{Laine2020ModularPF}.
Note that we do not assume a known pose for the individual images in the real distribution; however, we assume a distribution on the poses from which viewpoints are sampled. 
Images of the generated textured mesh are then rendered from views sampled from the distribution, which are then critiqued by 2D convolutional discriminators. 
We use two discriminators: the first, like conventional discriminators, considers whether a single image comes from the real or generated distributions; the second considers whether a set of rendered views from a generated textured shape is consistent across the shape. 
Since we do not have access to multiple views of the same sample from the real distribution,  we consider multiple patches from a single real image sample. 
For generated images, we then consider multiple patches from different views as input to the patch-consistency discriminator.
As patches coming from the same view have a consistent style, the generated patches across views are also encouraged to have a matching style. 
Operating at patch level is important since for small patches it is harder to distinguish if the patches are coming from the same or from different viewpoints. 
Fig.~\ref{fig:method_patchdisc}(b) shows the effect of this patch-consistency discriminator, as considering only single views independently can lead to artifacts where the front and back of a shape are textured inconsistently, while the patch consistency across views leads to a globally consistent textured output.
Both discriminators use the architecture of the original StyleGAN2 discriminators and use adaptive discriminator augmentation~\cite{karras2020training}.

\subsection{Implementation Details}
We use a hierarchy of $n=6$ quad-meshes with number of faces as (24576, 6144, 1536, 384, 96, 24) from finest to coarsest resolution respectively. Pooling layers use a \textit{mean} operation for aggregating features. During training, each mesh is rendered at a resolution of $512\times512$ across $4$ random viewpoints. The patch consistency discriminator uses patches cropped to a resolution of $64\times64$, with $4$ patches extracted from each generated viewpoint, yielding a total of $16$ patches as input. The generator uses an empirically determined weighting of 10:1 for losses coming from the image discriminator and the patch discriminator.

\noindent Our \OURS{} model is implemented using Pytorch and trained using Adam~\cite{kingma2014adam} with learning rates of $1\times10^{-4}$, $2\times10^{-3}$ and $1\times10^{-3}$ for the encoder, decoder and both discriminators respectively. 
We train on 2 NVIDIA A6000s for 70k iterations ($\sim 80$ hours) until convergence. We plan to open source our model, data and data-processing scripts.

\begin{figure}[b!]  
    \centering
    \includegraphics[width=\linewidth]{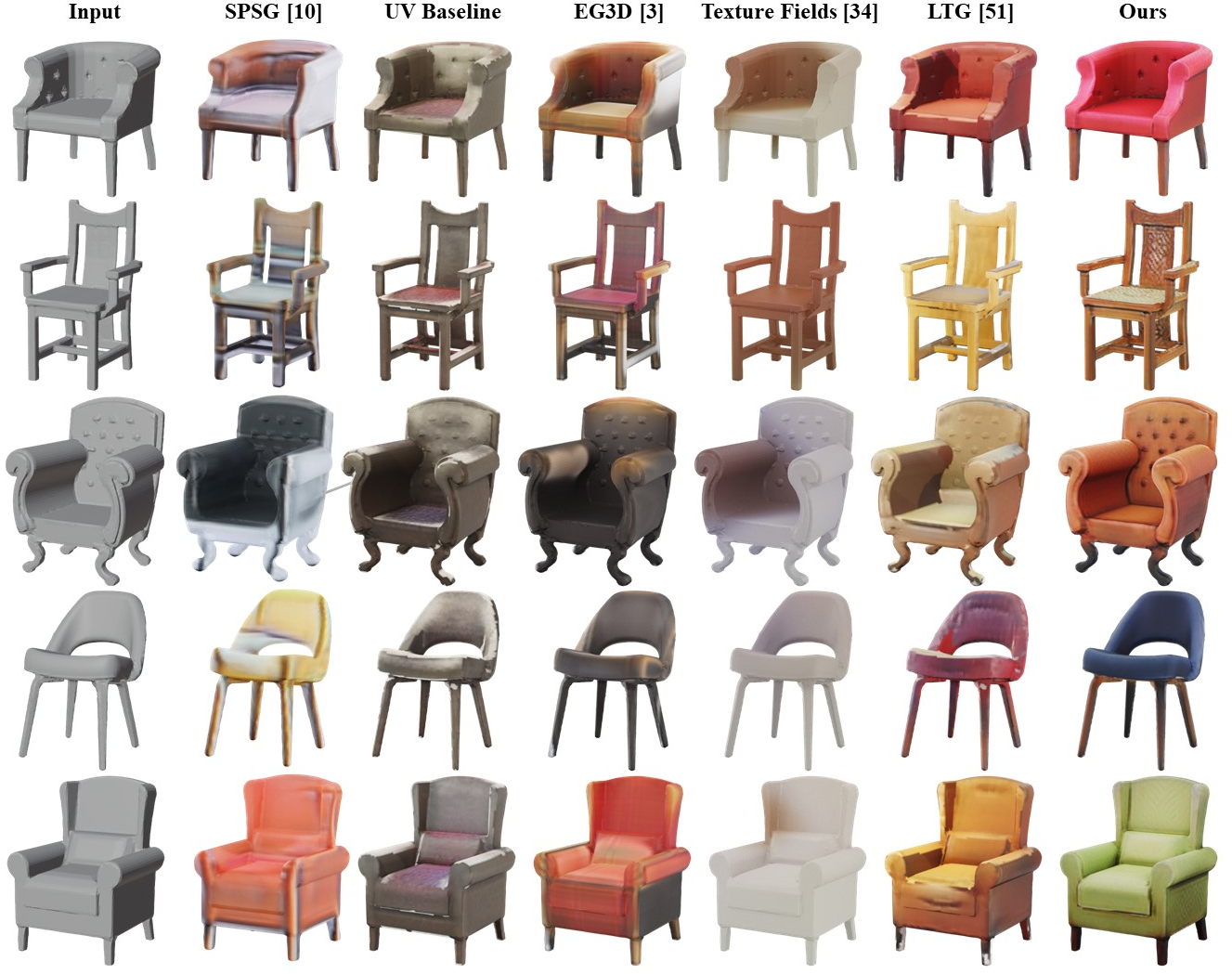}
    \vspace{-0.75cm}
    \caption{Qualitative results on ShapeNet chairs dataset trained with real images from the Photoshape dataset. While methods producing textures in 3D space like SPSG~\cite{dai2021spsg}, EG3D~\cite{ChanARXIV2021} and TextureFields~\cite{oechsle2019texture} produce blurry textures, UV based methods like LTG~\cite{yu2021learning} show artifacts at UV seams, specially for non-convex shapes like chairs. By operating on the surface, our method can generate realistic and detailed textures.}
    \label{fig:results_chair}
\end{figure}

\section{Experiments}
\label{sec:results}

\noindent \textbf{Data.}
We evaluate our method on 3D shape geometry from the `chair' and `car' categories of the ShapeNet dataset~\cite{shapenet2015}. 
For chairs, we use 5,097 object meshes split into 4,097 train and 1,000 test shapes, and 15,586 images from the Photoshape dataset~\cite{park2018photoshape} which were collected from image search engines. 
For cars, we use 1,256 cars split into 956 train and 300 test shapes.
We use 18,991 real images from the CompCars dataset~\cite{yang2015large} and use an off-the-shelf segmentation model~\cite{kirillov2020pointrend} to obtain foreground-background segmentations.

\smallskip
\noindent \textbf{Evaluation Metrics.}
Our evaluation is based on common GAN image quality and diversity metrics.
Specifically, we use the Frechet Inception Distance (FID)~\cite{heusel2017gans} and Kernel Inception Distance (KID)~\cite{binkowski2018demystifying} for evaluating the generative models.
For each mesh, we render images of the textured shapes produced by each method at a resolution of $256\times 256$ from 4 random view points using 4 random latent codes, and evaluate these metrics against all available real images segmented from their background.
Note that we do not have ground truth textures available for the 3D shapes and specific style codes, thus, a classical reconstruction metric (e.g., an $\ell_1$ distance) is not applicable.

\begin{figure}[t]  
    \centering
    \includegraphics[width=\linewidth]{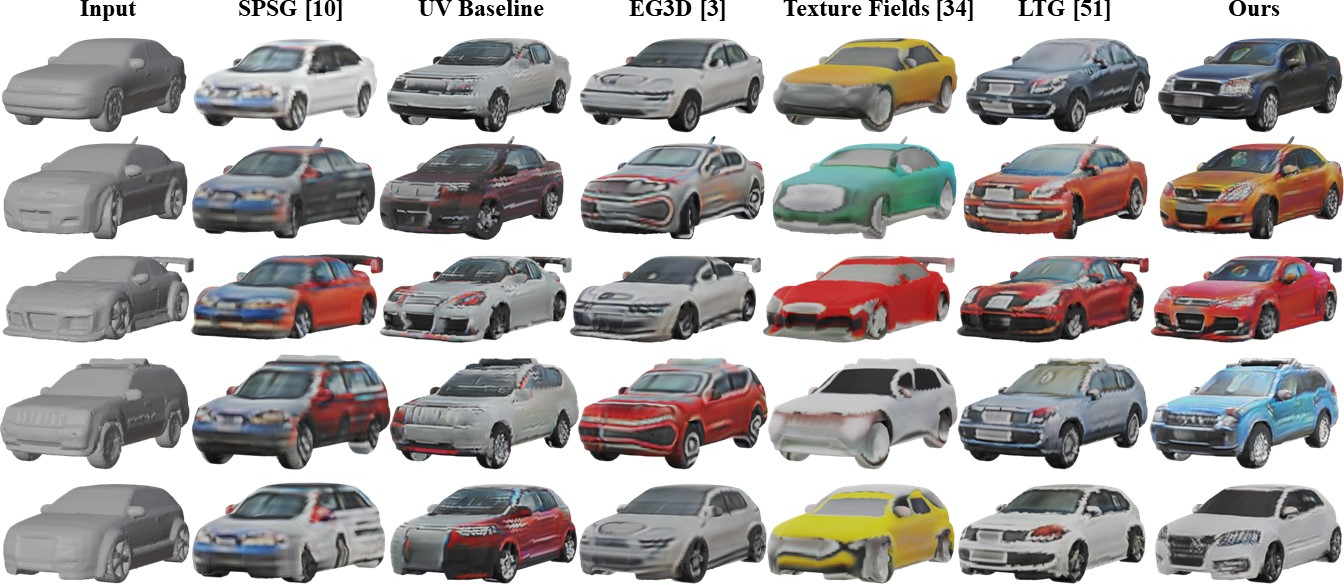}
    \caption{Qualitative results on ShapeNet cars trained with real images from the CompCars dataset. Our method can generate realistic and diverse cars textures on varying shape geometries like sedans, sportscars, SUVs and hatchbacks.}
    \label{fig:results_car}
\end{figure}

\setlength{\tabcolsep}{4pt}
\begin{table}[b]
    \begin{center}
    \caption{Comparison against state-of-the-art texture generation approaches on ShapeNet chairs and cars learned on real-world 2D images.}
    \label{table:comparison}
    \begin{tabular}{|l|l|r|r|r|r|}
    \hline
    \multirow{2}{*}{Method} & \multirow{2}{*}{Parameterization} & \multicolumn{2}{c|}{Chairs} & \multicolumn{2}{c|}{Cars} \\
    \cline{3-6}
     & & KID$\times 10^{-2}$$\downarrow$ & FID$\downarrow$ & KID$\times 10^{-2}$$\downarrow$ & FID$\downarrow$ \\
    \hline
    Texture Fields~\cite{oechsle2019texture} & Global Implicit & $6.06$ & $85.01$ & $17.14$ & $177.15$\\
    SPSG~\cite{dai2021spsg} & Sparse 3D Grid & $5.13$ & $65.36$ & $9.59$ & $110.65$ \\
    UV Baseline & UV & $2.46$ & $38.98$ & $5.77$ & $73.63$\\
    LTG~\cite{yu2021learning} & UV & $2.39$ & $37.50$ & $5.72$ & $70.06$ \\
    EG3D~\cite{ChanARXIV2021} & Tri-plane Implicit & $2.15$ & $36.45$ & $5.95$ & $83.11$\\
    \hline
    Ours & 4-RoSy Field & $\mathbf{1.54}$ & $\mathbf{26.17}$ & $\mathbf{4.97}$ & $\mathbf{59.55}$\\ 
    \hline
    \end{tabular}
    \end{center}
\end{table}

\smallskip
\noindent \textbf{Comparison against state of the art.}
Tab.~\ref{table:comparison} shows a comparison to state-of-the-art methods for texture generation on 3D shape meshes.
We compare with Texture Fields~\cite{oechsle2019texture}, which generates textures as implicit fields around the object, Yu et al.~\cite{yu2021learning} which learns texture generation in the UV parameterization space, and a modified version of EG3D~\cite{ChanARXIV2021} such that it uses a hybrid explicit-implicit tri-plane 3D representation to predict textures for a given mesh conditioned on its geometry.
Additionally, we compare with the voxel-based 3D color generation of SPSG~\cite{dai2021spsg}; since this was originally formulated for scan completion, we adopt its differentiable rendering to our encoder-decoder architecture using 3D convolutions instead of FaceConvs, with sparse 3D convolutions in the final decoder layer.
Finally, we also compare to a UV-based baseline which takes our network architecture with 2D convolutions to learn directly in UV space rather than on a 4-RoSy field.
In contrast to these alternatives, our approach generates textures directly on the mesh surface, maintaining local geometric reasoning which leads to more compelling appearance generation for both chair and car meshes (see Fig.~\ref{fig:results_chair} and \ref{fig:results_car}).
The network architectures for our method and baselines are detailed in the supplementary.

\begin{figure}[h]  
    \centering
    \includegraphics[width=\linewidth]{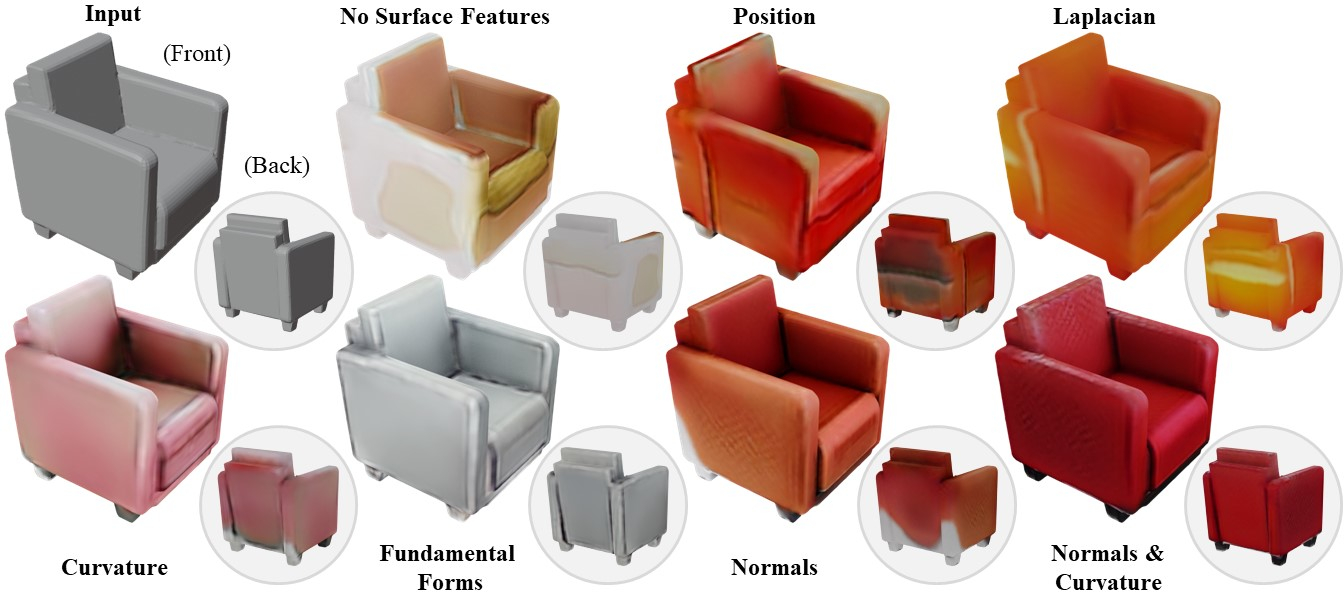}
    \caption{Effect of different input surface features. Using no surface features (i.e. no surface encoder, only the decoder) produces poor textures since the decoder has limited understanding of the shape. 3D location based features such as position and Laplacian suffer from the inability to effectively align texture patterns with geometric ones. Curvature and fundamental forms introduce spurious line effects due to strong correlation with curvature. Surface normal are quite effective, and are further stabilised when used along with curvature.}
    \label{fig:ablation_features}
\end{figure}

\smallskip
\noindent \textbf{Which surface features are the most informative for texture generation?} 
We evaluate a variety of different local surface features used as input to the encoder network, see Tab.~\ref{table:ablations}.
In particular, we consider a case \textit{`None'}, where we do not use a surface feature encoder, thus, the surface StyleGAN generator is reasoning only via the mesh neighborhood structure, a case where we input the 3D \textit{position} of the face centroid, and the cases with local geometric neighborhood characterizations using \textit{Laplacian}, \textit{curvature}, \textit{discrete fundamental forms}, and \textit{normals} as input features.
We find that using surface features help significantly over using no features (\textit{`None'}).
Further, features dependent on surface geometry like curvature, fundamental forms perform better than positional features like absolute 3D position and Laplacian, with a combination of normals and curvature providing the most informative surface descriptor for our texture generation (see Fig.~\ref{fig:ablation_features}).

\setlength{\tabcolsep}{4pt}
\begin{table}[h]
    \begin{center}
    \caption{Ablations on geometric input features, mesh and image resolution, number of views, and the patch-consistency discriminator for our method on ShapeNet chairs.}
    \vspace{-0.35cm}
    \label{table:ablations}
    \resizebox{\linewidth}{!}{
    \begin{tabular}{|l|l|l|l|l|r|r|}
    \hline
    Input Feature & Mesh-Res & Render-Res & \# views & Patch D & KID$\times 10^{-2}$$\downarrow$ & FID$\downarrow$ \\
    \hline
    \cellcolor{LightBlue}None & 24K & 512 & 4 & \checkmark & 2.53 & 37.95 \\
    \cellcolor{LightBlue}Position & 24K & 512 & 4 & \checkmark & 2.10 & 34.45 \\
    \cellcolor{LightBlue}Laplacian & 24K & 512 & 4 & \checkmark & 2.05 & 34.18\\
    \cellcolor{LightBlue}Curvatures & 24K & 512 & 4 & \checkmark & 1.79 & 29.86\\
    \cellcolor{LightBlue}Fundamental Forms & 24K & 512 & 4 & \checkmark & 1.80 & 30.91\\
    \cellcolor{LightBlue}Normals & 24K & 512 & 4 & \checkmark & 1.68 & 27.73\\
    Normals + Curvature & \cellcolor{LightBlue}6K & 512 & 4 & \checkmark & 2.01 & 33.95\\
    Normals + Curvature & 24K & \cellcolor{LightBlue}64 & 4 & \checkmark & 2.35 & 39.32 \\
    Normals + Curvature & 24K & \cellcolor{LightBlue}128 & 4 & \checkmark & 1.93 & 32.22\\
    Normals + Curvature & 24K & \cellcolor{LightBlue}256 & 4 & \checkmark & 1.54 & 26.99 \\
    Normals + Curvature & 24K & 512 & \cellcolor{LightBlue}1 & \checkmark & 1.67 & 27.83\\
    Normals + Curvature & 24K & 512 & \cellcolor{LightBlue}2 & \checkmark & 1.56 & 26.95\\
    Normals + Curvature & 24K & 512 & 4 & \cellcolor{LightBlue}\text{\sffamily X} & 1.64 & 27.22\\ 
    \hline
    Normals + Curvature & 24K & 512 & 4 & \checkmark &  \textbf{1.54} & \textbf{26.17}\\ 
    \hline
    \end{tabular}
    }
    \end{center}
\end{table}

\begin{figure}  
    \centering
    \includegraphics[width=\linewidth]{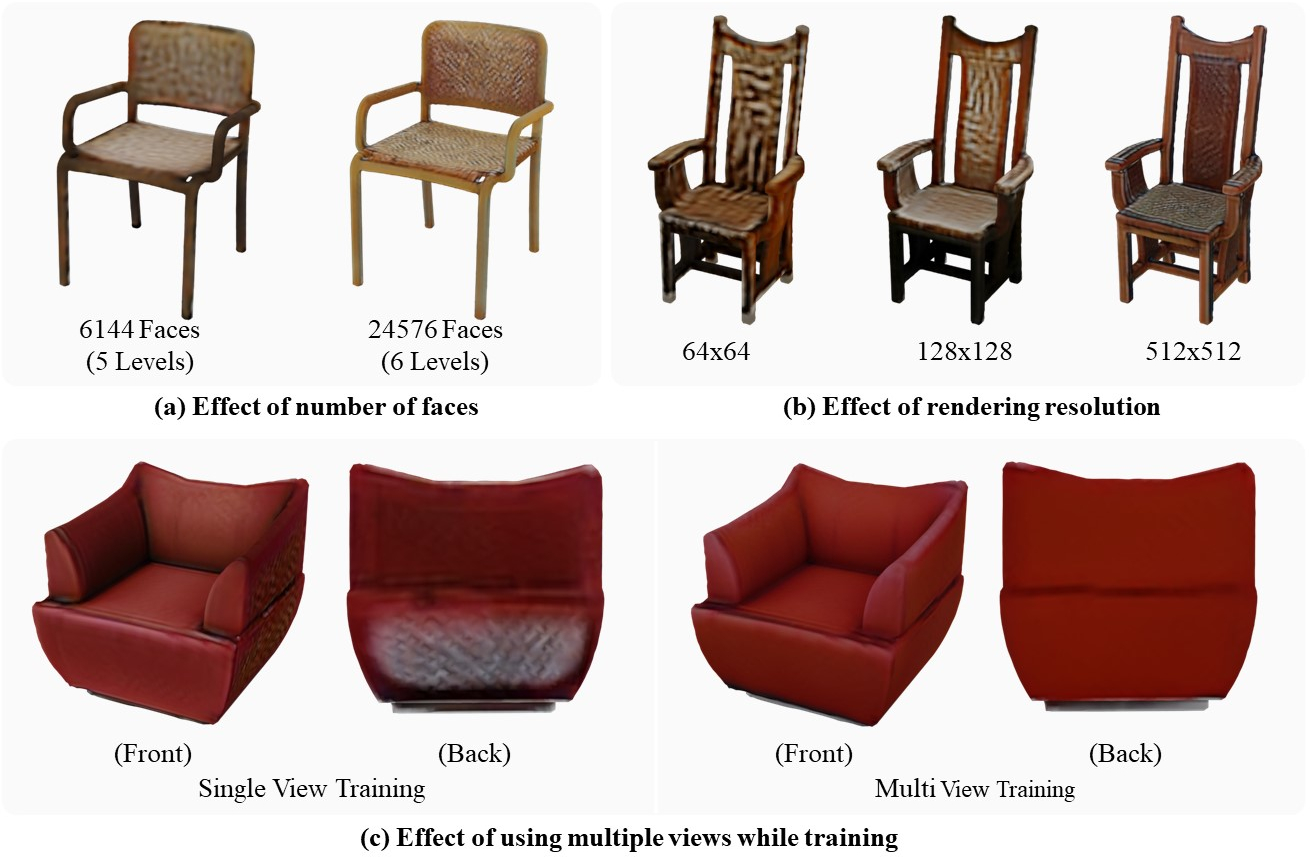}
    \caption{(a) Increased mesh resolution enables synthesis of higher quality texture. (b) Higher rendering resolution during training helps synthesize more details. (c) Using a single viewpoint per mesh for discrimination can introduce artifacts in the texture across different regions of the mesh. In contrast, using multiple views instead encourages more coherent textures across views.
    }
    \label{fig:ablation_facecount_renderres_views}
\end{figure}

\begin{figure} 
\centering
\includegraphics[width=\linewidth]{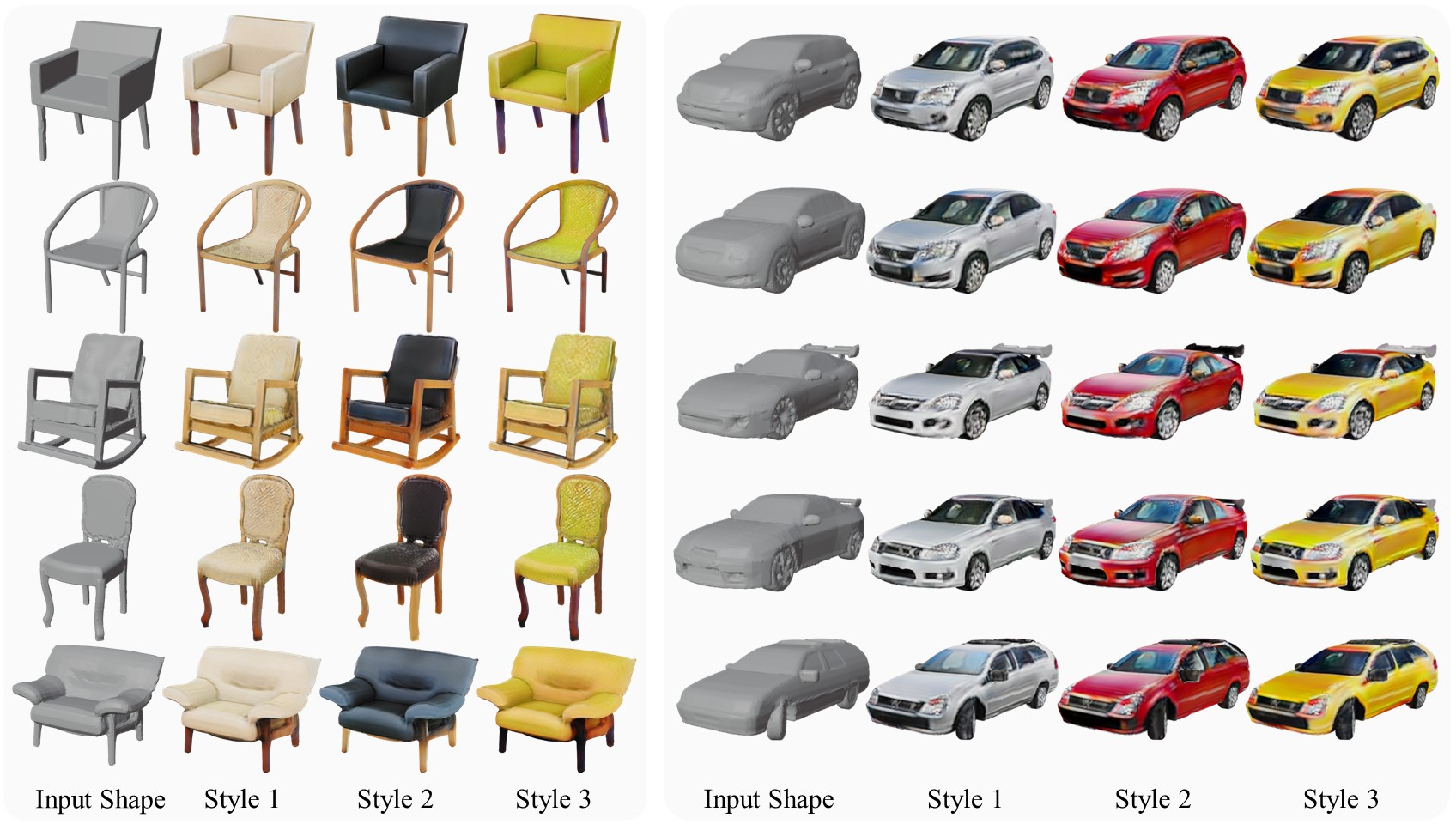}
\caption{The latent texture space is consistent across different shape geometry, such that the same latent code gives a similar texture style for different geometries.}
\label{fig:consistency}
\end{figure}

\smallskip
\noindent \textbf{What is the impact of the patch-consistency discriminator?}
When a patch consistency discriminator is not used (see Fig.~\ref{fig:method_patchdisc}), our method can end up generating textures that might look valid from distinct view points, but as a whole incorporate different styles.
Using a discriminator that considers patch consistency across multiple different views enables more globally coherent texture generation (Fig.~\ref{fig:method_patchdisc}, right), also reflected in improved KID and FID scores (Tab.~\ref{table:ablations}, last two lines).

\smallskip
\noindent \textbf{What is the effect of the mesh resolution on the quality?} 
We compare our base method with 6 hierarchy levels with number of faces as (24576, 6144, 1536, 384, 96, 24) against our method with 5 hierarchy levels with number of faces as (6144, 1536, 384, 96, 24).
As seen in Tab.~\ref{table:ablations} and Fig.~\ref{fig:ablation_facecount_renderres_views}(a), the increased mesh resolution helps to produce higher-quality fine-scale details, resulting in notably improved performance.
Even higher resolutions on our setup were prohibitive in memory consumption.
%

\smallskip
\noindent \textbf{How does the rendered view resolution affect results?} 
Tab.~\ref{table:ablations} and Fig.~\ref{fig:ablation_facecount_renderres_views}(b) show the effect of several different rendering resolutions during training: 64, 128, 256, and 512.
Increasing the rendering resolution results in improved quality enabling more effective generation of details.

\smallskip
\noindent \textbf{Does rendering from multiple viewpoints during training help?}
We consider a varying number of viewpoints rendered per object during each optimization step while training in Tab.~\ref{table:ablations}, using 1, 2, and 4 views.
We see that using more views continues to help when increasing from 1 to 2 to 4.
Specifically, we see that using multiple views helps avoid artifacts that appear in the mesh across vastly different viewpoints as shown in Fig.~\ref{fig:ablation_facecount_renderres_views}(c). Note that the multi-view setting also allows patch consistency discriminator to generate consistent textures.
We use 4 views during training since increasing the number of views has a decreasing marginal benefit as views will become more redundant. 

\begin{figure}[h]  
    \centering
    \includegraphics[width=\linewidth]{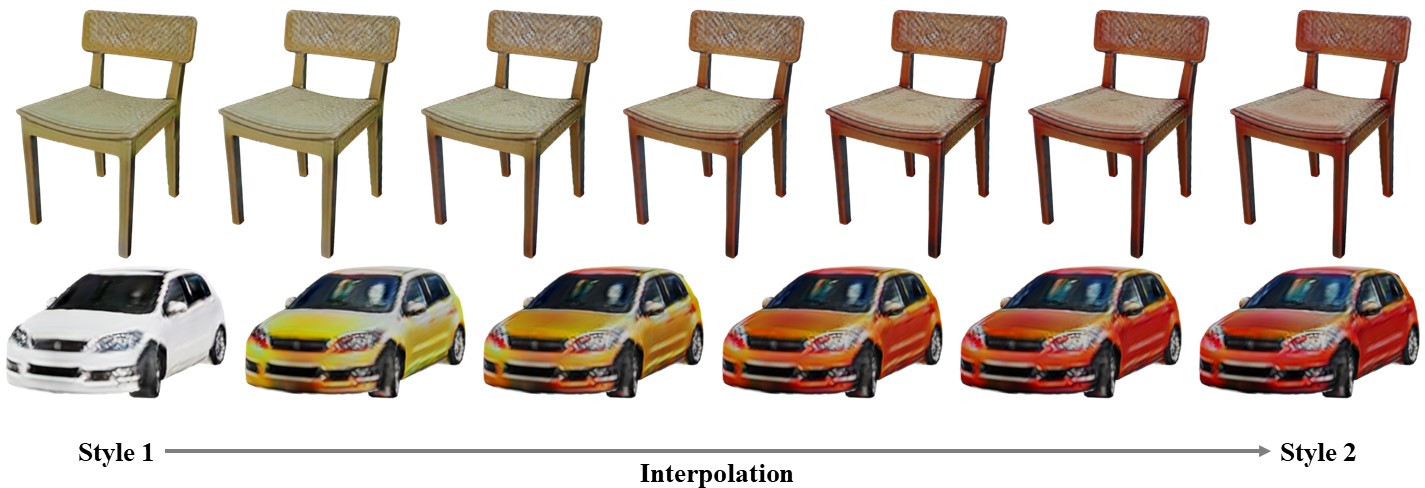}
    \caption{The texture latent space learned by our method produces smoothly-varying valid textures when traversing across the latent space for a fixed shape.}
    \vspace{-0.50cm}
    \label{fig:latent}
\end{figure}

\smallskip
\noindent \textbf{Learned texture latent space behavior.}
For a fixed shape, our learned latent space is well behaved with a smooth interpolation yielding valid textures, as shown in Fig.~\ref{fig:latent}.
Furthermore, the learned latent space is consistent in style across different shapes, i.e. the same code represents a similar style across shapes (Fig.~\ref{fig:consistency}), and can be used, for example, in style transfer applications.

\paragraph{Limitations.}
While our approach takes a promising step towards texture generation on shape collections, some limitations remain.
Since our method uses a real image distribution that comes with lighting effects, and our method does not factor out lighting, the textures learned by our method can have lighting effects baked in. 
Further, the resolution of the texture generated by our method is limited by the number of faces used at the highest level of the 4-RoSy parameterization, whereas learned implicit functions or explicit subdivison at these highest-level faces could potentially capture even higher texture resolutions.

\section{Conclusion}
We have introduced \textit{\OURS{}}, a new approach to generate textures on mesh surfaces from distinct collections of 3D shape geometry and 2D image collections, i.e., without requiring any correspondences between 2D and 3D or any explicit 3D color supervision.
Our texture generation approach operates directly on a given mesh surface and synthesizes high-quality, coherent textures.
In our experiments we show that the 4-RoSy parameterization in combination with face convolutions using geometric features as input outperforms the state-of-the-art methods both quantitatively, as well qualitatively.
We believe that \textit{\OURS{}} is an important step in 3D content creation through automatic texture generation of 3D objects which can be used in standard computer graphics pipelines.

\clearpage

\smallskip
\noindent \textbf{Acknowledgements.}
This work was supported by the Bavarian State Ministry of Science and the Arts coordinated by the Bavarian Research Institute for Digital Transformation (bidt), a TUM-IAS Rudolf M{\"o}{\ss}bauer Fellowship, an NVidia Professorship Award, the ERC Starting Grant Scan2CAD (804724), and the German Research Foundation (DFG) Grant Making Machine Learning on Static and Dynamic 3D Data Practical. Apple was not involved in the evaluations and implementation of the code. Further, we thank the authors of LTG~\cite{yu2021learning} for assistance with their code and data.

\bibliographystyle{splncs04}
\bibliography{egbib}

\clearpage
\begin{appendix}

\section{Network Architecture}
\begin{figure}[h]  
    \centering
    \includegraphics[width=\linewidth]{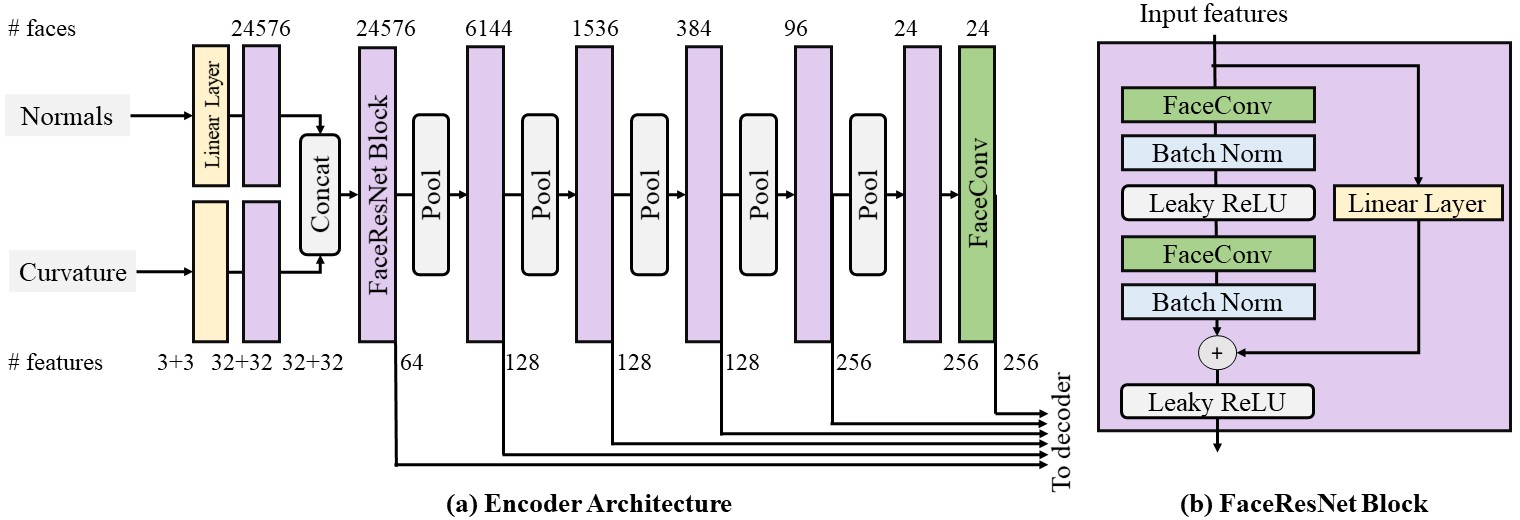}
    \caption{
    Encoder architecture.
    (a) The encoder takes in face normals and curvature at the finest resolution of the hierarchy and extracts features using FaceResNet blocks (b).
    Features are extracted at all levels of the hierarchy using inter hierarchy pooling and are passed on to the decoder in a U-Net style with skip connections (see Fig.~\ref{fig:supp_arch_decoder}).
    (b) A FaceResNet block is a ResNet block that uses FaceConv instead of Conv2D, and therefore can operate on surface of the mesh.
    }
    \label{fig:supp_arch_encoder}
\end{figure}

\begin{figure}[h]  
    \centering
    \includegraphics[width=\linewidth]{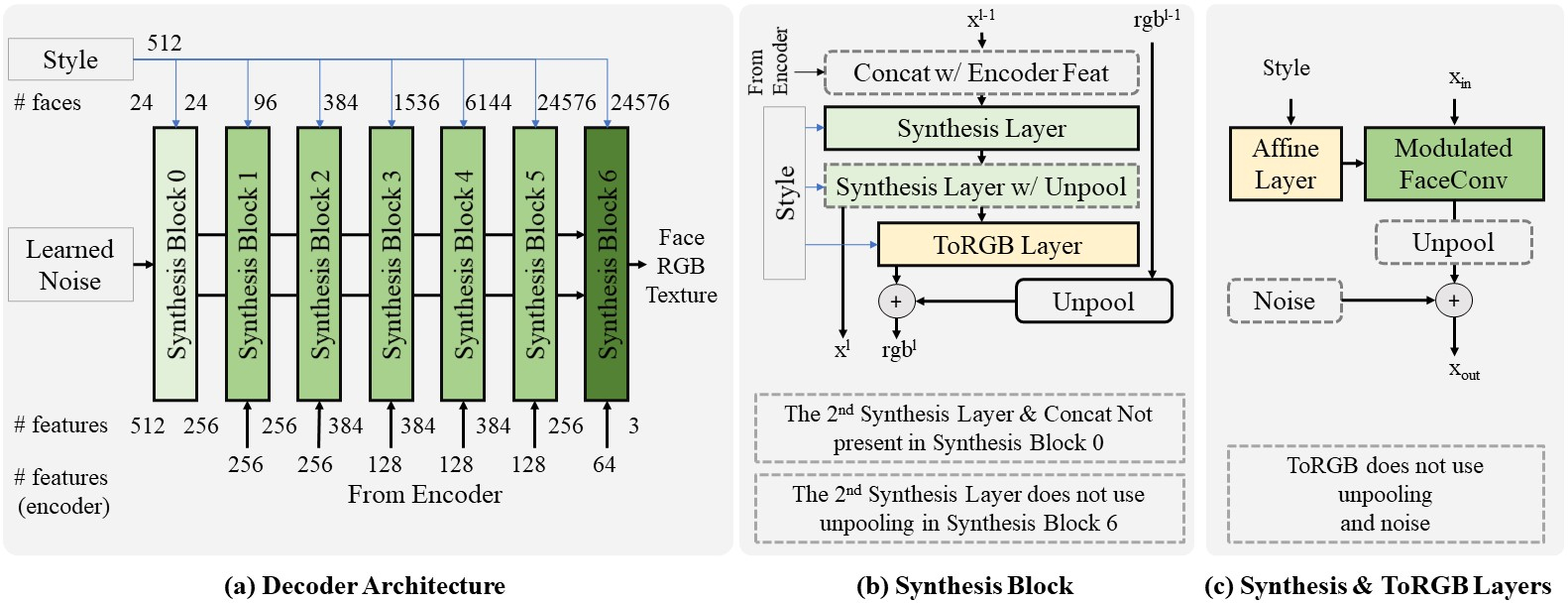}
    \caption{
    Decoder architecture.
    (a) The decoder is inspired by the StyleGAN2 generator, but rather than operating on a 2D image hierarchy, it operates on quad mesh hierarchy.
    A learned noise over a cube, which is always the coarsest resolution with 24 faces in the hierarchy, is upsampled to a specific shape through a series of synthesis blocks that take in surface features coming from the encode and the style codes.
    (b) A synthesis block concatenates features coming from the encoder with the previous synthesis blocks generated features, and passes them through synthesis layers, one of which unpools them to a finer level in the hierarchy.
    Features generated at the current level are also decoded to an RGB texture and added to the unpooled RGB texture coming from previous layer.
    (c) Synthesis layer applies a FaceConv with weights modulated by a style code to the input features, and optionally unpools and adds noise to it.
    }
    \label{fig:supp_arch_decoder}
\end{figure}

A detailed description of the network for our method can be found in Fig.~\ref{fig:supp_arch_encoder} and \ref{fig:supp_arch_decoder}.
Fig.~\ref{fig:supp_arch_encoder} details the encoder, while Fig.~\ref{fig:supp_arch_decoder} displays the StyleGAN2 inspired decoder.
Both are based on our 4-RoSy parametrization and the FaceConv and pooling operations presented in the main paper.

\medskip
\noindent \textbf{Singularities.}
Singularities are vertices on a quad mesh that do not have a valancy of 4.
Our method uses zero-padding on faces with singularity vertices to enforce a fixed size neighborhood.
Quad meshes parameterized by quadriflow have few singularities.
Specifically, we get $0.89\%$ vertices with singularities for Chair category, and $1.95\%$ for Car category.
This is comparable to the proportion of pixel locations that need padding in a $256\times256$ image ($1.56\%$).

\begin{figure}[h]  
    \centering
    \includegraphics[width=\linewidth]{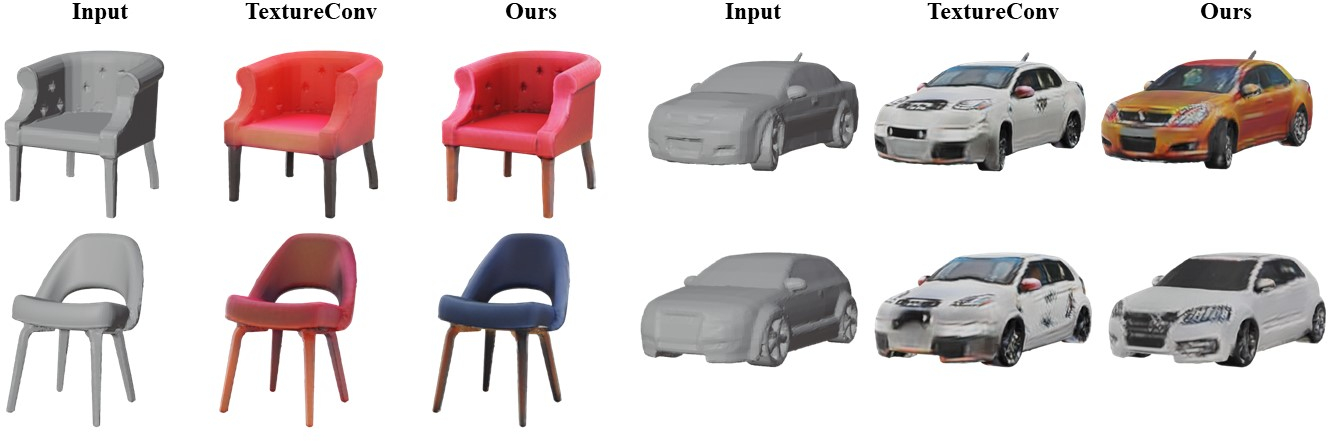}
    \vspace{-0.75cm}
    \caption{Comparison with a modified version of our network using texture convolutions from TextureNet~\cite{huang2019texturenet}.
    }
    \vspace{-0.25cm}
    \label{fig:supp_textureconv}
\end{figure}

\medskip
\noindent \textbf{FaceConvs.}
As described in the main paper, our approach uses FaceConvs as operator on a 4-RoSy surface.
A similar 4-RoSy parameterization has been used in TextureNet~\cite{huang2019texturenet} for the segmentation of point clouds.
Instead of our FaceConvs, which use Cartesian ordering to resolve the 4 way ambiguity, TextureNet introduced TextureConvs a 4-RoSy surface convolutional operator.
In Table.~\ref{table:supp_tconv}, we modify our proposed method and replace the FaceConvs with these TextureConvs.
We observe that while TextureConvs work reasonably well for the chair category, it struggles with the placement of headlights and the front grill for the car category (Fig.~\ref{fig:supp_textureconv}). 

\setlength{\tabcolsep}{4pt}
\begin{table}[h!]
    \vspace{-0.2cm}
    \begin{center}
    \caption{Comparison against a modified version of our network that uses TextureConv~\cite{huang2019texturenet} instead of FaceConvs on ShapeNet chairs and cars learned on real-world 2D images.
    Our proposed FaceConvs lead to significantly better geometry-aware texture synthesis, especially, on the car dataset (see Fig.~\ref{fig:supp_textureconv}).
    }
    \label{table:supp_tconv}
    \begin{tabular}{|l|r|r|r|r|}
    \hline
    \multirow{2}{*}{Method} & \multicolumn{2}{c|}{Chairs} & \multicolumn{2}{c|}{Cars} \\
    \cline{2-5}
     & KID$\times 10^{-2}$$\downarrow$ & FID$\downarrow$ & KID$\times 10^{-2}$$\downarrow$ & FID$\downarrow$ \\
    \hline
    TexureConv~\cite{huang2019texturenet} & $1.88$ & $31.67$ & $6.13$ & $80.10$\\
    Ours & $\mathbf{1.54}$ & $\mathbf{26.17}$ & $\mathbf{4.97}$ & $\mathbf{59.55}$\\ 
    \hline
    \end{tabular}
    \end{center}
    \vspace{-0.5cm}
\end{table}

\newpage

\section{Baseline Methods}
\begin{figure}[h!]  
    \centering
    \includegraphics[width=1.0\linewidth]{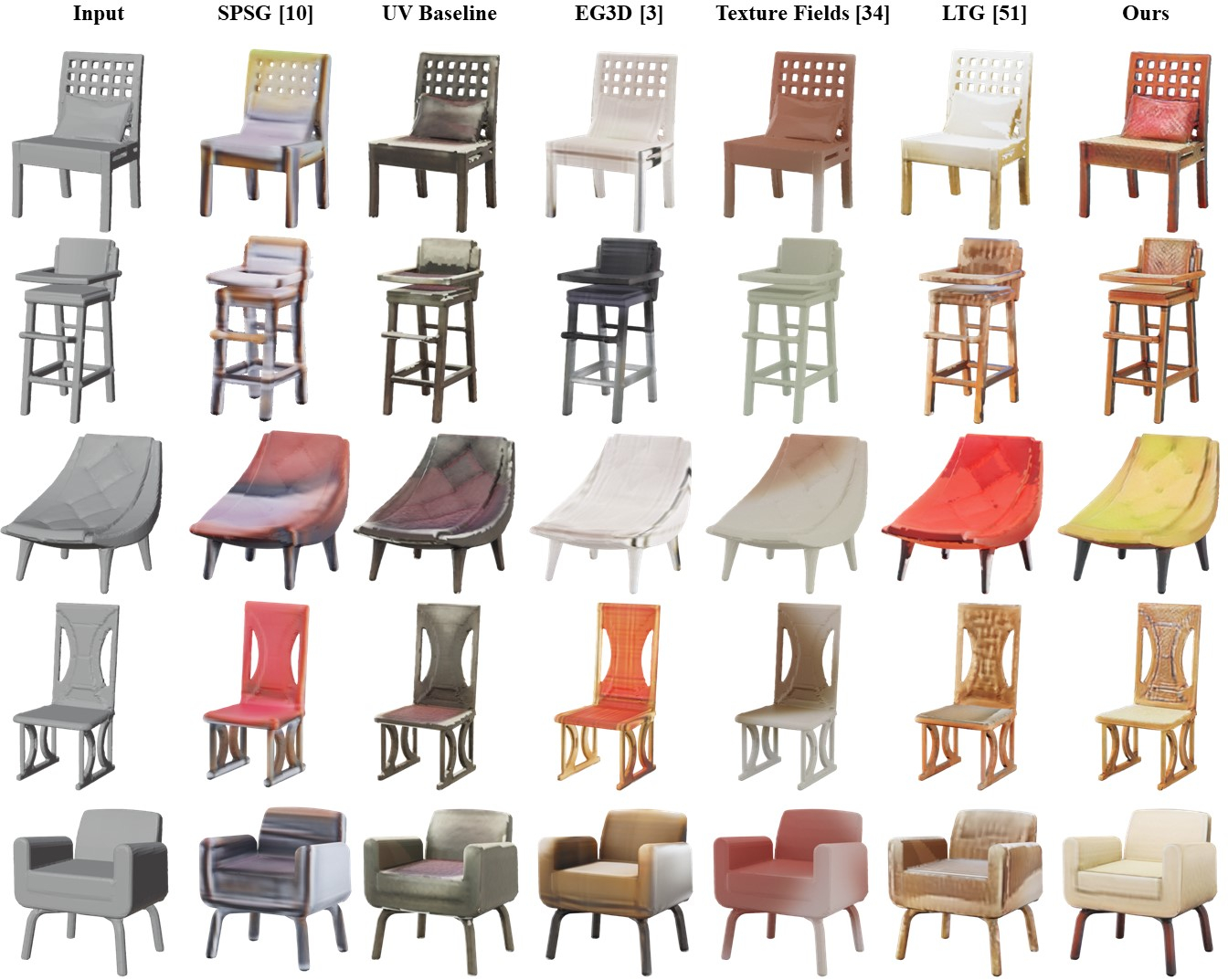}
    \vspace{-0.25cm}
    \caption{Qualitative results on ShapeNet chairs dataset trained with real images from the Photoshape dataset}
    \label{fig:supp_results_chairs}
\end{figure}

\begin{figure}[h!]  
    \centering
    \includegraphics[width=1.0\linewidth]{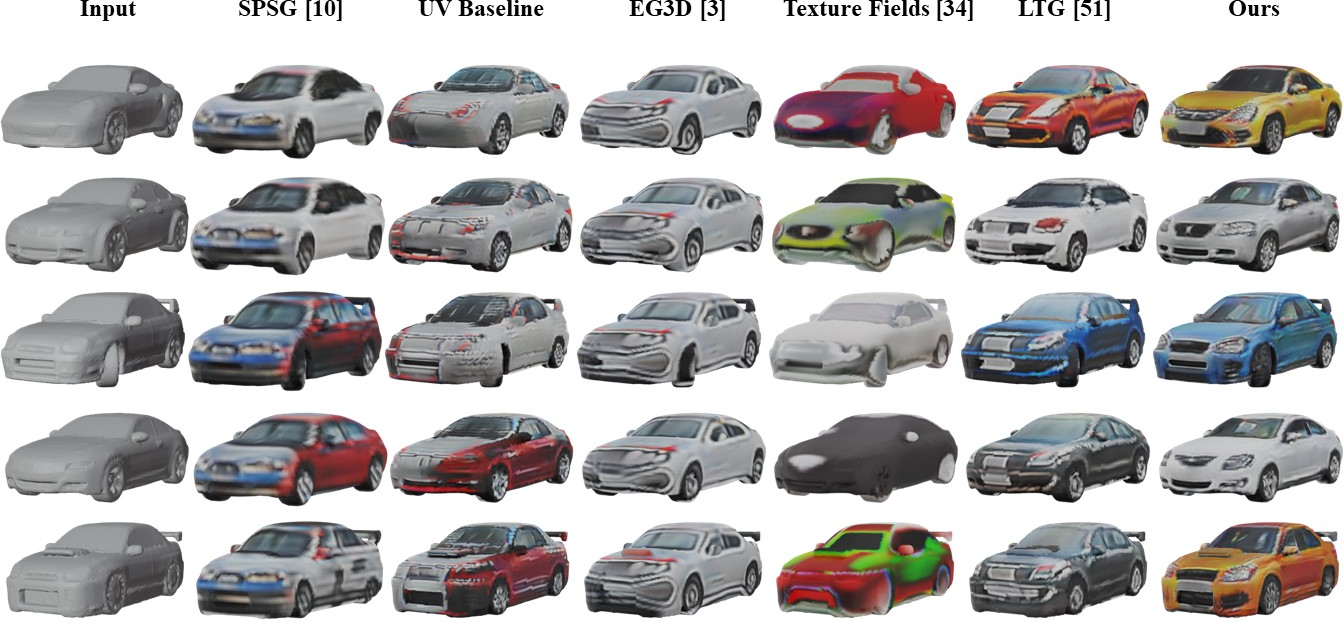}
    \vspace{-0.25cm}
    \caption{Results on ShapeNet cars trained with real images from the CompCars dataset.}
    \label{fig:supp_results_cars}
\end{figure}

We describe the experimental setup for the various baseline comparisons with state-of-the-art texture generation, along with an additional quantitative comparison on the Generated Image Quality Assessment (GIQA) metric~\cite{gu2020giqa} in Tab.~\ref{table:supp_comparison} and additional qualitative comparisons in Fig.~\ref{fig:supp_results_chairs} and \ref{fig:supp_results_cars}.
The methods differ mainly in their parametrization as discussed below.

\setlength{\tabcolsep}{4pt}
\begin{table}[h!]
    \vspace{-0.2cm}
    \begin{center}
    \caption{Comparison against state-of-the-art texture generation approaches on ShapeNet chairs and cars learned on real-world 2D images.}
    \label{table:supp_comparison}
    \begin{tabular}{|l|l|r|r|}
    \hline
    \multirow{2}{*}{Method} & \multirow{2}{*}{Parameterization} & \multicolumn{2}{c|}{GIQA$\times 10^{-2}$$\uparrow$} \\
    \cline{3-4}
     & & Chairs & Cars \\
    \hline
    Texture Fields~\cite{oechsle2019texture} & Global Implicit & $6.29$ & $5.14$ \\
    SPSG~\cite{dai2021spsg} & Sparse 3D Grid & $6.38$ & $7.19$ \\
    UV Baseline & UV & $7.29$ & $7.84$\\
    LTG~\cite{yu2021learning} & UV & $7.39$ & $7.90$ \\
    EG3D~\cite{ChanARXIV2021} & Tri-plane Implicit & $7.58$ & $7.85$\\
    \hline
    Ours & 4-RoSy Field & $\mathbf{7.73}$ & $\mathbf{7.99}$\\ 
    \hline
    \end{tabular}
    \end{center}
    \vspace{-0.5cm}
\end{table}

\begin{figure}[h]  
    \centering
    \vspace{-0.35cm}
    \includegraphics[width=\linewidth]{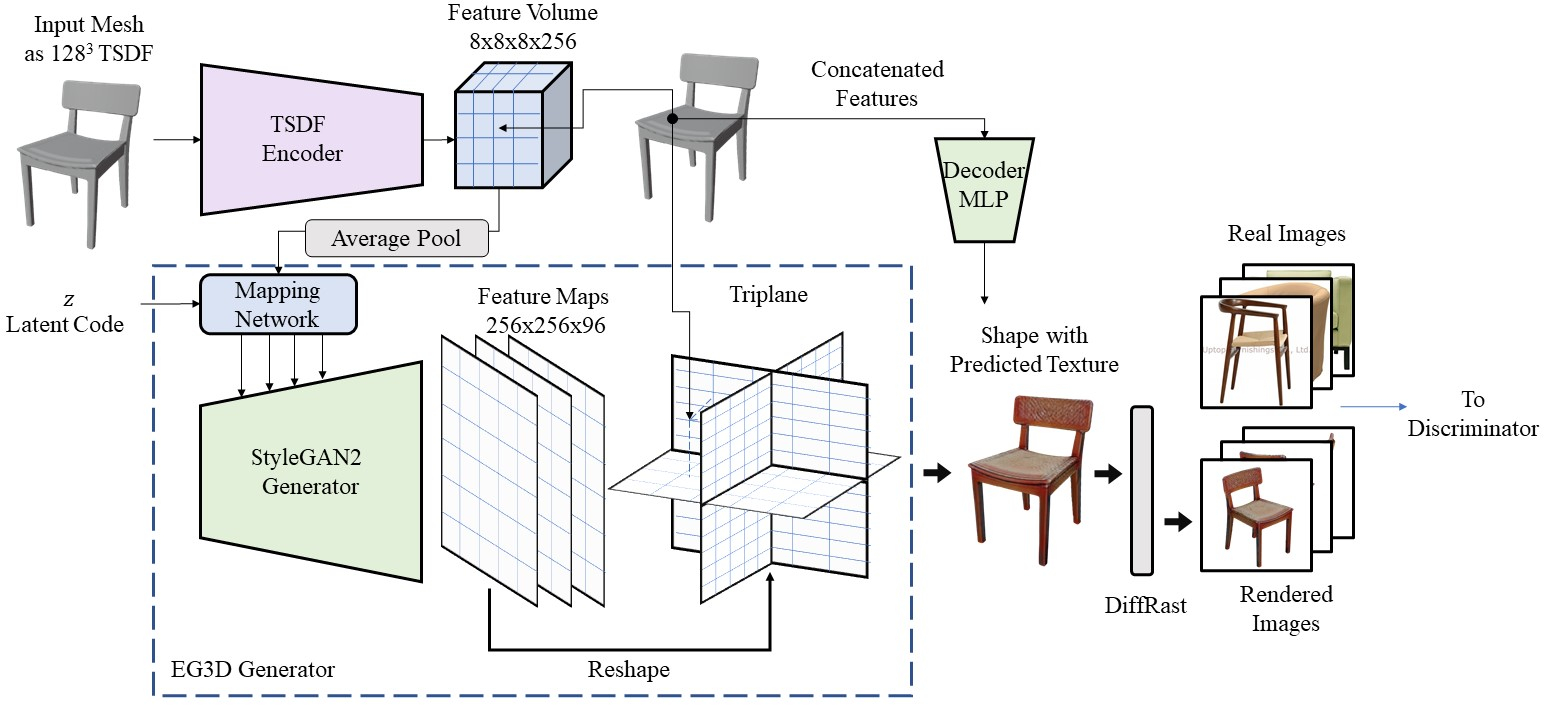}
    \caption{EG3D~\cite{ChanARXIV2021} inspired baseline architecture. A StyleGAN2 generator outputs a triplane representation with style conditioned on a mesh code. The input mesh represented as a TSDF grid is additionally encoded into an $8^3$ feature volume. For points on the mesh surface, features are sampled from the Triplane and 3D feature grid, concatenated, and decoded via an MLP to get face colors. The resulting mesh with face colors is differentiably rendered and critiqued by a discriminator.}
    \label{fig:supp_eg3d}
    \vspace{-0.35cm}
\end{figure}

\noindent \textbf{TextureFields.}
For TextureFields~\cite{oechsle2019texture}, we use the official code and configuration of its GAN variant.
We found that training with purely real-world images made the network unstable, so we used a mix (with a probability $p=0.5$) of real images and synthetic renders with ShapeNet textures.
\medskip
\noindent \textbf{SPSG.}
For the SPSG~\cite{dai2021spsg} inspired baseline, we use the exact same architecture as ours (Fig.~\ref{fig:supp_arch_encoder} and \ref{fig:supp_arch_decoder}), except that instead of surface, the networks now operate in a 3D grid.
A TSDF grid at the finest resolution of $128^3$ is input to the encoder, with features extracted and pooled using 3D ResNet blocks (ResNet block with Conv3D) and trilinear downsampling operators.
The decoder uses modulated Conv3Ds instead of modulated FaceConvs and unpooling is performed using trilinear upsampling of features.
The last synthesis block uses sparse convolutions because of memory constraints.
The decoder outputs a 3D grid of RGB colors instead of per face RGB colors.
This color grid is rendered to an image with the shape's TSDF using SPSG's TSDF differentiable rendering.
\medskip
\noindent \textbf{UV-Space.}
For the UV baseline, we again use broadly the same architecture as ours.
Here, instead of operating on surface using 4-RoSy parameterization, we operate on the surface using UV parameterization.
Specifically, we compute the UV maps for the shapes in a fashion similar to LTG~\cite{yu2021learning}, i.e. using 6 views (top, bottom, left, right, front, back) around the object.
Input to the encoder are the normal and curvature atlas maps.
Features are extracted using vanilla 2D ResNet blocks at multiple resolutions and passed to the decoder.
The decoder is a regular 2D StyleGAN conditioned (through concatenation like ours) on features coming from encoder.
It predicts texture atlases which are mapped to the shape during differentiable rendering.
This pipeline differs from LTG as it does not use SPADE-IN blocks for conditioning on silhouettes but instead conditions on surface features extracted via the encoder.
Further, this baseline synthesises a single texture atlas unlike the multiple texture atlases in LTG.
\medskip
\noindent \textbf{EG-3D.}
Finally, the EG3D~\cite{ChanARXIV2021} inspired baseline architecture is shown in Fig.~\ref{fig:supp_eg3d}.
Here, given an input mesh and its 3D TSDF representation, a StyleGAN2 network generates a triplane representation, while a 3D TSDF encoder encodes a 3D feature grid.
An MLP decoder is then used to query face colors point in space, based on the features projected on the triplane and the feature on the grid at the query point.
The mesh with it's face colors is then differentiably rendered and critiqued through a discriminator. 

\section{Discussion \& Outlook}

\begin{figure}[h]  
    \centering
    \vspace{-0.35cm}
    \includegraphics[width=\linewidth]{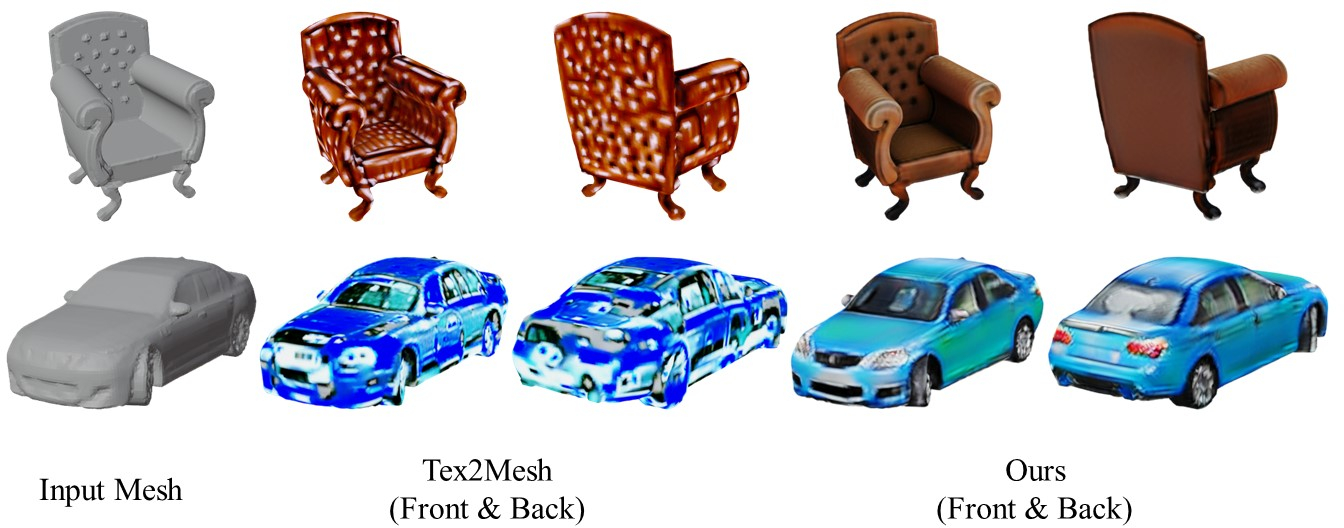}
    \caption{Comparison with Text2Mesh~\cite{michel2021text2mesh} with queries ``brown chair" (top) and ``blue sedan car" (bottom).}
    \vspace{-0.35cm}
    \label{fig:supp_text2mesh}
\end{figure}

Our method learns to texture 3D objects from in-the-wild image datasets.
It exhibits consistent global and local structural details and can also be used for text-based texture synthesis.
To this end, we adapted the Text2Mesh~\cite{michel2021text2mesh} framework to take advantage of our texture model.
Specifically, we optimize the latent code passed to our pretrained generator using an evolutionary algorithm such that the CLIP~\cite{radford2021learning} scores between query and the renders are maximized.
In Fig.~\ref{fig:supp_text2mesh}, we show a comparison to the original Text2Mesh approach, where we only optimize for the colors on the surface of the mesh.
Note that we disable the geometry optimization for this experiment.
While Text2Mesh gives good textures when the queries specify a small scale texture description like ``brick" or ``cactus'', it fails to synthesize textures in a semantically consistent way for broader queries like ``brown chair'' or ``blue sedan car''.
For instance, in the case of the car mesh, Text2Mesh synthesizes smaller images of cars on the surface of the car mesh.
In contrast, our method generates semantically consistent textures, also on a higher abstraction level.

\begin{figure}[h]  
    \centering
    \includegraphics[width=\linewidth]{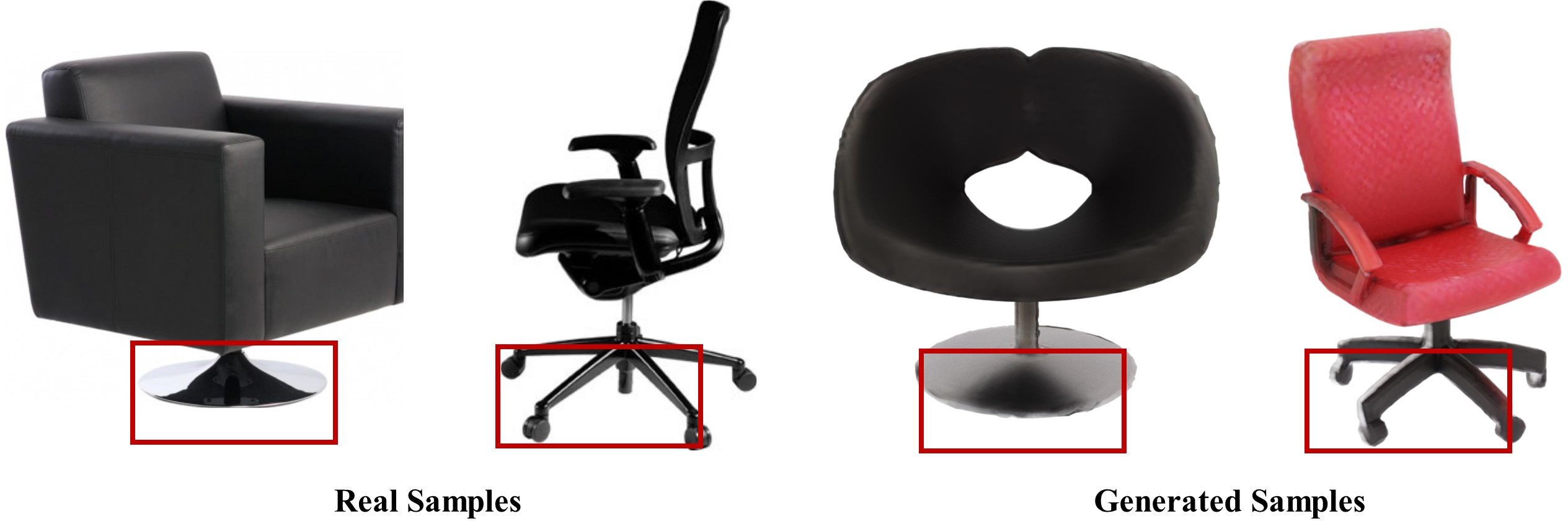}
    \caption{Since our method does not model illumination, the textures produced by our method can end up replicating the lighting effects found in the training images.}
    \label{fig:supp_limitations}
\end{figure}

While we already see a wide applicability of our method, there are limitations that we want to address in future work.
As we learn from real world data, we also capture lighting effects, e.g., shadows or specular highlights in our texture.
These `baked-in' effects might look reasonable from one view-point, but view-dependent effects like specular highlights should not be synthesized in the texture since they are implausible from other view-points (see Fig.~\ref{fig:supp_limitations}).
Therefore, additional effort has to be invested to disentangle these effects from the actual diffuse texture.
In addition, we think that combining our texture estimation approach that estimates per face colors, with local texture MLPs similar to IF-Nets or ConvOcc (which could predict a color for each point on a face) is an interesting avenue for future research.

\end{appendix}

\end{document}